\documentclass[letterpaper]{article} 
\usepackage{aaai2026}  
\usepackage{times}  
\usepackage{helvet}  
\usepackage{courier}  
\usepackage[hyphens]{url}  
\usepackage{graphicx} 
\usepackage{natbib}  
\usepackage{caption} 
\usepackage{algorithm}
\usepackage{algorithmic}
\usepackage{colortbl}

\usepackage{subcaption}
\usepackage{multirow}
\usepackage{comment}
\usepackage{soul}
\usepackage{amsmath}
\usepackage{amssymb}
\usepackage{mathtools}
\usepackage{amsthm}
\usepackage{booktabs}
\usepackage{dsfont}

\theoremstyle{plain}
\newtheorem{theorem}{Theorem}[section]

\theoremstyle{definition}
\newtheorem{definition}[theorem]{Definition}

\theoremstyle{remark}

\usepackage{pifont}
\definecolor{darkgreen}{RGB}{0,128,0}
\newcommand{\tick}{\textcolor{darkgreen}{\ding{51}}}  
\newcommand{\cross}{\textcolor{red}{\ding{55}}}  

\usepackage{newfloat}
\usepackage{listings}
\DeclareCaptionStyle{ruled}{labelfont=normalfont,labelsep=colon,strut=off} 
\floatstyle{ruled}
\newfloat{listing}{tb}{lst}{}
\floatname{listing}{Listing}
\title{Revisiting (Un)Fairness in Recourse by Minimizing Worst-Case Social Burden\footnote{\color{red} \textbf{This manuscript is an extended version of our work published at AAAI 2026.
Please cite the conference version when referencing this material.}}}
\author{
    Ainhize Barrainkua\textsuperscript{\rm 1}, Giovanni De Toni\textsuperscript{\rm 2}, Jose A. Lozano\textsuperscript{\rm 1,3}, Novi Quadrianto\textsuperscript{\rm 1,4}
}
\affiliations{
    \textsuperscript{\rm 1}Basque Center for Applied Mathematics, Bilbao, Spain\\
    \textsuperscript{\rm 2}Fondazione Bruno Kessler, Trento, Italy\\
    \textsuperscript{\rm 3}University of the Basque Country, Donostia - San Sebastian, Spain \\
    \textsuperscript{\rm 4}Predictive Analytics Lab, University of Sussex, Brighton, UK \\
    abarrainkua@bcamath.org,
    gdetoni@fbk.eu,
    ja.lozano@ehu.eus,
    n.quadrianto@sussex.ac.uk
}

\begin{document}

\maketitle

\begin{abstract}
Machine learning based predictions are increasingly used in sensitive decision-making applications that directly affect our lives.
This has led to extensive research into ensuring the fairness of classifiers.
Beyond just fair classification, emerging legislation now mandates that when a classifier delivers a negative decision, it must also offer actionable steps an individual can take to reverse that outcome. 
This concept is known as \textit{algorithmic recourse}.
Nevertheless, many researchers have expressed concerns about the fairness guarantees within the recourse process itself. 
In this work, we provide a holistic theoretical characterization of unfairness in algorithmic recourse, formally linking fairness guarantees in recourse and classification, and highlighting limitations of the standard equal cost paradigm. We then introduce a novel fairness framework based on \textit{social burden}, along with a practical algorithm (\texttt{MISOB}), broadly applicable under real-world conditions. Empirical results on real-world datasets show that \texttt{MISOB} reduces the social burden across all groups without compromising overall classifier accuracy.
\end{abstract}

\begin{links}
    \link{Code}{https://github.com/abarrainkua/MISOB}
    \link{Extended version}{https://arxiv.org/abs/2509.04128}
\end{links}

\begin{figure*}[t]
    \centering
    \begin{subfigure}[b]{0.59\textwidth} 
        \centering
        \includegraphics[width=\columnwidth]{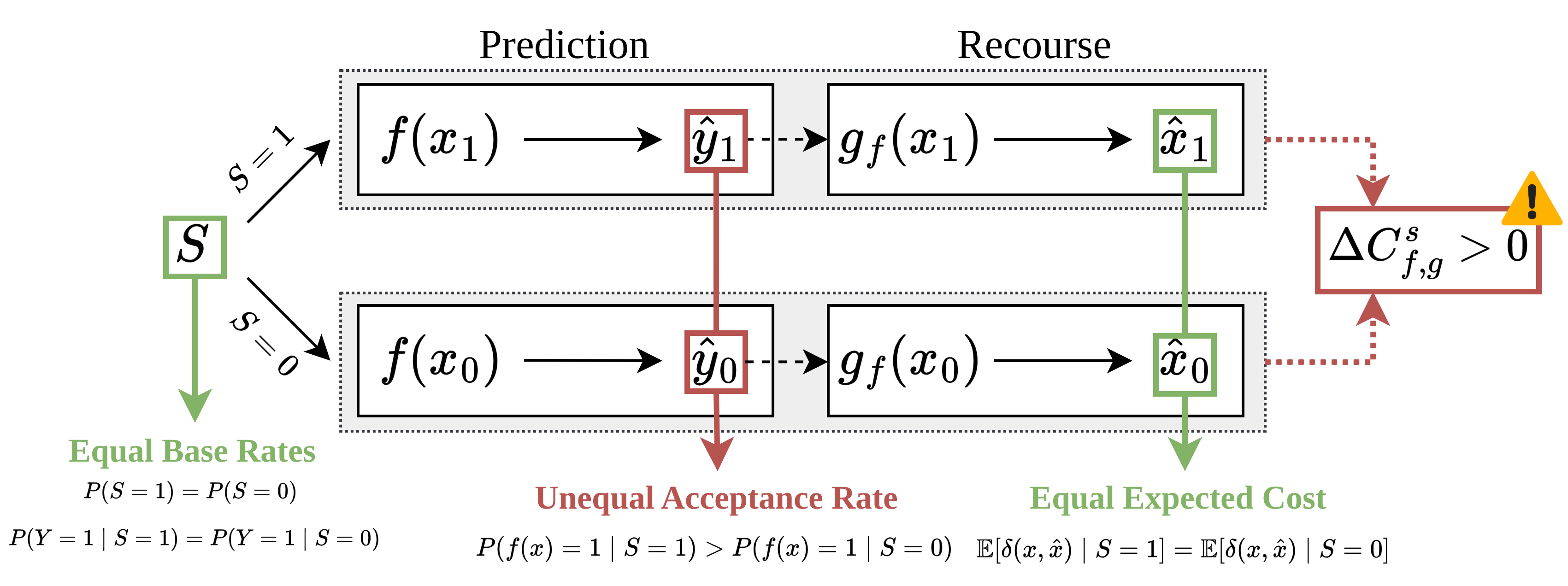}
        \caption{The separate treatment of fairness for predictive and recourse elements.}
        \label{fig:eq_cost_diagram}
    \end{subfigure}
    \hfill
    \begin{subfigure}[b]{0.40\textwidth} 
        \centering
        \includegraphics[width=\columnwidth]{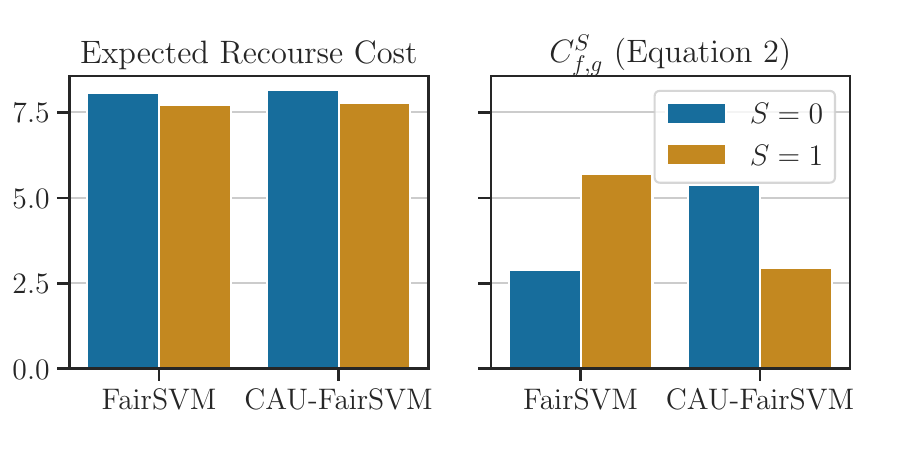} 
        \caption{Holistic metrics expose large disparities.}
        \label{fig:eq_cost_limit}
    \end{subfigure}
    \caption{\textbf{Conventional metrics to measure fairness in recourse hide disparities.} (a) Given a population balanced by sensitive group membership and ground‑truth outcomes (\textit{e.g.,} true ability to repay a loan), treating prediction and recourse fairness separately, as in the majority of prior work, can mask disparities: if positive classifications are unevenly distributed, one group bears recourse burdens more frequently even when per‑instance recourse costs are equal; misclassification (\textit{e.g.,} false negatives) further amplifies this effect. (b) Conventional cost‑parity metrics suggest balanced recourse costs across groups $S \in \{0,1\}$ (left). Our metric (Eqn.~\ref{eq:cost_group}) reveals substantial disparity once group differences in decision and error rates are taken into account, even with ``fair'' classifiers, FairSVM \cite{gupta2019equalizing} and CAU-FairSVM \cite{von2022fairness} (right). 
    }
    \label{fig:unfairness_recourse}
\end{figure*}

\section{Introduction}
The ubiquity of automated decision-making systems has considerably reshaped the landscape of how individuals access crucial resources and are granted important social and economic opportunities, ranging from securing credit to receiving essential public services \cite{chouldechova2018case,purificato2023use,perdomo2025difficult}. 
In recent years, growing concerns have emerged regarding the fairness of these systems, as well as their societal impact \citep{grgichlaca2018human,jobin2019global,raji2022fallacy}. Notably, instances have been documented in which algorithmic decision support systems produced discriminatory outcomes in domains such as welfare distribution \citep{heikkila2021dutchscandal} and criminal justice \citep{angwin2022machine}.
Moreover, in high-stakes scenarios, it is imperative that, whenever a model yields a negative decision, it also provides the affected individual with clear and actionable \textit{recourse} \cite{karimi2022survey}.
Unfortunately, affected individuals usually lack transparency and effective means to contest undesired decisions (\textit{e.g.,} bail denial) or take meaningful steps to improve their situation\footnote{For example, independent audits have shown fairness concerns and limited contestability options in nationwide-deployed systems, such as \textit{COMPAS} for recidivism prediction in the United States \cite{angwin2022machine} and \textit{VioGén} for gender-based violence risk assessment in Spain \cite{eticas2022viogen}.}.
For this reason, the ``\textit{right to recourse}'' is slowly emerging as a legal requirement in several legislations, \textit{e.g.,} in the European GDPR and the Artificial Intelligence (AI) Act \cite{act2021proposal}. 

Pursuing a given recourse often demands considerable effort from the affected individual.
As noted by \citet{ustun2019actionable}, the average effort to achieve recourse can vary significantly across sensitive groups, raising important concerns about fairness in recourse.
This disparity may arise when (i) the recourse itself \textit{differs} depending on \textit{group membership}, or when (ii) the recourse is formally identical, but the real-world \textit{effort} required to carry it out \textit{varies} in practice.
For example, consider a method that offers suggestions to individuals denied a loan. It might recommend \textit{"Increase your monthly income by \$500"} to someone from a privileged group, while advising \textit{"Maintain stable employment for 24 more months"} to someone from an underprivileged background.
The first recommendation is often more viable for individuals with access to high-paying jobs or stable employment opportunities.
In contrast, maintaining long-term employment can be substantially more challenging for individuals in precarious jobs or members of low-income and otherwise underprivileged communities.
Further, consider an algorithm suggesting that all rejected students retake standardized tests. It may impose minimal burden on those who can afford test preparation, tutoring, and multiple attempts, while students coming from low social and economic backgrounds often face barriers, including limited resources and time constraints, due to many socio-economic factors \cite{rodriguez2020socio}.
Thus, what appears to be group-aware or uniform recommendations can translate into very different levels of effort, reinforcing structural inequalities.

In light of these concerns, it is paramount to develop metrics to reliably quantify and monitor the fairness of recourse recommendations, along with practical methods for improving these metrics.
Existing research has primarily examined unfairness in recourse by analyzing the costs imposed on individuals receiving negative classifications (\textit{i.e.,}``reject''), either at the individual or group level~\cite{von2022fairness,ehyaei2023robustness,kavouras2023fairness,bell2024fairness,yetukuri2024providing}.
However, we argue that fairness in recourse depends not only on \textit{what} type of recourse is provided but also on \textit{who} is more likely to receive the negative decision in the first place.
As an illustration, consider two groups of people, $S=0$ and $S=1$, having the same number of individuals per group.
If one of the groups (\textit{e.g.,} $S=1$) has a higher acceptance rate than the other group, many more individuals from $S=0$ receive a ``reject'' decision.
Widely utilized fairness-in-recourse metric -- \textit{equal expected cost (of negatively predicted individuals)} -- will not capture the inherent unfairness in the system as we can balance the cost of recourse per individual across two groups (\textit{e.g.,} $10/100 = 100/1000 =0.1$).
However, individuals in group $S=0$ are disproportionately classified as negative ($1000$ individuals vs. $100$) and therefore, a higher number of instances from $S=0$ bear the cost of recourse.

We emphasize that the above illustrated problem is amplified when we consider the \textit{ground-truth labels}, \textit{e.g.,} an individual’s latent (and unobserved) ability to repay a loan instead of just the classifier's acceptance outcomes.
Individuals may be wrongly classified and burdened with corrective actions, despite the error stemming from the decision support system itself.
This misallocation of responsibility is troubling, particularly as error rates often differ systematically across sensitive groups, a well-documented challenge in algorithmic fairness~\cite{mehrabi2022surveyfairness}.
All the issues mentioned above stem from the fact that current popular fairness-in-recourse paradigms do not correctly capture fairness issues in the whole system, which includes both the classifier's components and the recourse's components.

This paper presents a unified framework for assessing fairness in algorithmic recourse across the full decision pipeline, from initial classification through recourse recommendation (see Fig.~\ref{fig:eq_cost_diagram} for an overview).
Our framework yields formal connections between \textit{prediction fairness} and \textit{recourse fairness}, showing how disparities in classification performance can propagate to recourse recommendations, and vice versa.
Further, by leveraging these insights, we introduce a lightweight training method, \texttt{MISOB} (MInimax SOcial Burden), that empirically improves fairness throughout the pipeline, \textit{without requiring access to sensitive attributes}, in real-world datasets.
Lastly, we argue for moving beyond simple group-gap metrics and developing richer measures that more fully describe fairness guarantees.

\paragraph{Our contributions.} Summarizing, we (i) theoretically characterize the sources of unfairness in algorithmic recourse, establishing formal connections between fairness guarantees in recourse and classification (Section~\ref{sec:theo_analysis}), (ii) provide evidence of the limitations of approaches based on the standard equal cost paradigm (Section~\ref{subsec:classdec}), (iii) introduce a novel framework for fair recourse (Algorithm~\ref{alg:misob}) based on the concept of \textit{social burden} (Section~\ref{sec:miso}), and (iv) assess its performance on real-world datasets (Section~\ref{sec:experiments}). 

\begin{table*}[t]
\setlength{\tabcolsep}{12pt}
\begin{center}
\caption{
\textbf{Summary of the existing works on fairness in recourse.}
We report the proposed fairness metric, whether it considers the classifier ($f(x)$) the recourse method ($g_f(x)$) or the ground truth outcomes ($Y$), and the approach used to enhance fairness; specifically, whether it is agnostic to the classifier, the recourse method, or the sensitive attributes ($\mathcal{S}$).
A ``-'' indicates works that propose only evaluation metrics without introducing any algorithm or methodology improving fairness.
}
\label{tab:relworks}
\begin{sc}
\scalebox{0.8}{%
\begin{tabular}{lcccccc}
\toprule
                                                 & \multicolumn{3}{c}{Fairness Metric}  & \multicolumn{3}{c}{Approach agnostic to}                     \\
                                                 & $f(x)$  & $g_f(x)$ & $Y$  & $f(x)$ & $g_f(x)$ & $\mathcal{S}$ \\ \midrule
\citet{gupta2019equalizing}                        & \cross       &    \tick     & \cross      & \tick           & \cross            & \cross        \\
\citet{von2022fairness}                            & \cross       &    \tick     & \cross      & \tick           & \cross            & \cross        \\
\citet{raimondi2022equality}                      & \tick         &    \tick    & \cross       & \textbf{-}      & \textbf{-}        & \textbf{-}    \\
\citet{kuratomi2022measuring}                     & \tick         &   \tick     & \tick        & \textbf{-}      & \textbf{-}        & \textbf{-}    \\
\citet{kavouras2023fairness}                       & \cross       &    \tick     & \cross       & \textbf{-}      & \textbf{-}        & \textbf{-}    \\
\citet{bell2024fairness}                           & \cross        &   \tick     & \cross      & \tick           & \tick             & \cross        \\
\rowcolor{gray!20} \textbf{This Work} (Section~\ref{sec:theo_analysis} and \ref{sec:miso}) & \tick                 & \tick  &   \tick  & \tick           & \tick             & \tick         \\ \bottomrule
\end{tabular}%
}
\end{sc}
\end{center}
\end{table*}
\section{Preliminaries and Related Work}
\label{sec:background}

\paragraph{Notation and Problem Formulation.} Each decision subject (\textit{i.e.,} instance) is represented by the triplet\footnote{We denote random variables with capital letters, their realizations with lowercase letters, and sets with capital letters in \textit{italic}.} $(x, s, y)$, where $x \in \mathcal{X} \subseteq \mathbb{R}^d $ represents the non-sensitive attributes,  $s \in \mathcal{S} \subseteq \mathbb{R}^m$ the sensitive demographic attributes (\textit{e.g.,} race, gender, age), and  $y \in \mathcal{Y}$ the class label. For simplicity, we assume a binary classification task with $\mathcal{Y}= \{ 0,1 \}$ (\textit{e.g.,} whether the decision subject is capable of repaying the loan). However, the proposed framework is general and can be extended to multiclass settings.
The sensitive attributes $s$ can be multidimensional, enabling the characterization of individuals by multiple protected attributes and the analysis of intersectional groups (\textit{e.g.,} \textit{indigenous women under 30}) that reflect more nuanced and compounded socio-economic disadvantages.
Further, let $f: \mathcal{X} \rightarrow \mathcal{Y}$ be a \textit{binary} classifier whose goal is to assign a label in a set $\mathcal{Y}$ to instances in a set $\mathcal{X}$.
Let $\delta((x,s), x')$ denote the cost for an individual with features $(x,s)$ to change its non-sensitive profile to $x'$, \textit{e.g.,} $\ell_1$-norm \cite{karimi2022survey}. 
Given an instance that received a negative classification (\textit{i.e.,} $f(x) = 0$), we want to find the ``closest'' instance $x'$ achieving a positive classification by solving the following optimization problem: 
\begin{equation}
    \mathop{\mathrm{argmin}}_{x' \in \mathcal{X}} \; \delta((x,s), x') \quad \text{s.t.} \; f(x') \neq f(x)
\end{equation}
In the literature, $x'$ is also called \textit{counterfactual explanation} \cite{wachter2017counterfactual}.
Here, we assume the recommendation $x' = g_f(x)$ is generated by a recourse algorithm $g_f: \mathcal{X} \rightarrow \mathcal{X}$ tailored to the classifier $f$.
A growing body of work has proposed diverse strategies $g_f(x)$ to offer recourse to individuals assigned unfavorable predictions. 
These methods vary in several key aspects, including the type of base classifier considered, and actionability constraints.
Our discussion here is clearly limited; thus, please refer to \citet{karimi2022survey} for a comprehensive survey on the topic.
While providing recommendations is an important step toward accountability and transparency, significant fairness challenges remain, some stemming directly from the recourse process.
Indeed, recourse can exacerbate social segregation~\cite{gao2023impact}, and its cost may vary with user preferences~\cite{detoni2023synrec,detoni2024personalized} and sensitive attributes~\cite{yetukuri2024providing}, potentially deepening existing disparities.

\paragraph{Measuring Algorithmic Fairness.} Research on algorithmic fairness has primarily examined disparities in model predictions across population subgroups, focusing on the \textit{fairness of predictions themselves}.
Fairness definitions typically fall into two categories: \textit{individual-level}, which requires similar individuals to receive similar outcomes, and \textit{group-level}, which evaluates whether performance metrics (\textit{e.g.,} accuracy) are comparable across demographic groups.
Group metrics are widely used due to their computational simplicity, ease of integration into machine-learning pipelines, and ability to reveal structural disparities.
Two common examples are \textit{statistical parity} \cite{dwork2012fairness}, which measures differences in the probability of a positive prediction (\textit{i.e.,} acceptance rate) across groups regardless of true labels, and \textit{equality of opportunity} \cite{heidari2019moral}, which measures such differences only among individuals in the positive class (\textit{i.e.,} true positive rate). However, \textit{these metrics focus solely on the prediction stage and overlook the downstream consequences of model decisions}.

\paragraph{Measuring and Enhancing Fairness in  Recourse.} 

Fairness concerns in algorithmic recourse prompted the search for specific fairness criteria \cite{ustun2019actionable, gupta2019equalizing}.
Table~\ref{tab:relworks} summarizes the characteristics of the most relevant approaches that address fairness issues in recourse.
Many works proposed both group- and individual-level fairness metrics in this direction~\cite{von2022fairness, kavouras2023fairness, bell2024fairness} and extended the analysis to incorporate classifier predictions and standard predictive performance measures \cite{kuratomi2022measuring, raimondi2022equality}.
Several papers that introduced fairness metrics for evaluating algorithmic recourse have also proposed methods to improve fairness according to these metrics \cite{gupta2019equalizing, von2022fairness, bell2024fairness}.
For instance, \citet{von2022fairness} proposed a causality-based method that requires access to a structural causal model (SCM).
\citet{gupta2019equalizing} introduced an approach applicable only to recourse methods that map instances directly to the decision boundary, while \citet{bell2024fairness} addressed fairness under resource constraints, a setting closer to a ranking rather than a classification problem.

Most fairness metrics proposed in this context focus \textit{exclusively on the recourse stage}, considering only negatively classified instances and ignoring the classifier's overall behavior, such as group-specific acceptance and rejection rates.
Furthermore, they generally lack \textit{practical strategies} for improving fairness.
Lastly, none of these studies adopt an \textit{intersectional perspective}, and all assume access to sensitive attributes for all individuals \textit{at training time}, an assumption rarely satisfied in real-world applications.
Our work is conceptually aligned with \citet{kuratomi2022measuring} and \citet{raimondi2022equality}, but their formulation lacks a clear connection to existing fairness metrics in prediction.

\section{Revisiting (Un)Fairness in Recourse}
\label{sec:theo_analysis}

We now outline three key limitations of common fairness metrics in pipelines with recourse: (1) they ignore the classifier’s decision behavior, (2) they omit ground truth information, and (3) they focus solely on metric equalization.
We propose an alternative formulation that addresses these issues, offers a broader view of unfairness sources in recourse, and establishes explicit links between fairness in prediction and recourse.
Proofs are provided in Appendix \ref{app:proofs}.

\subsection{The Impact of the Classifier's Decisions}
\label{subsec:classdec}

Let us exemplify the first issue with an example of a binary classification task. 
Consider a population $X$ with a binary target variable $Y \in \{0,1\}$ and two equally represented sensitive groups $S \in \{0,1\}$, such that $P(S=0)=P(S=1)$ and $P(Y=1 \mid S=0) = P(Y=1 \mid S=1)$.
Let $f(x)$ be a classifier approximating the posterior $P(Y \mid X)$, and assume that under $f$ one group has a higher \textit{acceptance rate} (AR), $P(f(x)=1 \mid S=1) > P(f(x)=1 \mid S=0)$.
After recourse is provided to negatively classified instances, let us assume that the expected cost of the required changes is equal across groups
$\mathbb{E}[\delta((x, s), g_f(x)) \mid S=1] = \mathbb{E}[\delta((x,s), g_f(x)) \mid S=0]$.
According to standard fairness metrics, this situation would be considered fair. 
\textit{But is it truly fair?}
While the expected recourse cost is equal once offered, individuals in the group with lower AR are more likely to require recourse in the first place.
We argue that \textit{this disparity reflects an unfair situation}, which \textit{existing popular metrics fail to capture}.
We now introduce an alternative fairness metric that evaluates recourse costs across the entire population, rather than only among negatively classified instances.

\begin{definition}
    Let $f$ be a classifier and $g_f$ a recourse algorithm.
    Given a sensitive group $s\in\mathcal{S}$, let $\delta((x, S=s), g_f(x))$ be the cost of applying the recourse provided by $g_f$.
    Then, the expected recourse cost $C_{f,g}^{s}$ can be defined as:
    \begin{equation}
        \begin{aligned}
        \underbrace{\mathbb{E}[\delta((X,S=s),g_f(X))]}_{\text{Expected cost for instances with f(x)=0}} (1-\underbrace{P(f(X)=1 \mid S=s)}_{\text{Acceptance Rate (AR)}})
            \label{eq:cost_group}
        \end{aligned}
    \end{equation}
    where the expectation is over $P(X \mid S = s, f(X) = 0)$, and $P(f(X)=1 \mid S=s)$ indicates the \textit{acceptance rate} of the classifier for the given sensitive group. 
\end{definition}

Namely, Eqn.~\ref{eq:cost_group} arises since instances receiving a positive prediction, $\{x : f(x)=1, \;x \in X\}$, do not modify their non-sensitive features, implying no changes in recourse cost, $\delta((x,s), g_f(x)) = 0$. 
Fig.~\ref{fig:eq_cost_limit} shows how, when accounting for AR across sensitive groups, disparities in expected recourse costs become substantially larger.
Given the CAU-LIN dataset from \citet{von2022fairness}, we computed the empirical expected recourse cost by simply averaging only over negatively predicted instances (left), and by using Eqn.~\ref{eq:cost_group} (right), which also accounts for the AR of the base classifiers.
Indeed, Fig.~\ref{fig:eq_cost_limit} shows that both groups have similar costs under the original metric with FairSVM, but $S=1$ has the highest under the revised one, as a significantly larger proportion of this group receives negative predictions.
Therefore, to avoid masking underlying unfairness, we argue that the cost metric must account for \textit{acceptance rates} to accurately capture true recourse costs across groups.

\subsection{The Burden of False Negatives}

In Eqn.~\ref{eq:cost_group}, we measure the expected recourse cost while accounting for the classifier’s predictions, providing a more precise notion of fairness.
However, it focuses exclusively on the model’s predictions $f(x)$, without accounting for the true class label. As a result, it fails to distinguish between individuals who were incorrectly denied a positive outcome (\textit{i.e.,} truly positive but misclassified) and those who were correctly assigned a negative outcome.
In this section, we take a step further by focusing on minimizing the expected cost associated with positive instances that the classifier incorrectly predicts as negative. This next step naturally leads to the concept of social burden.\footnote{Some works on algorithmic recourse (\textit{e.g.,} \citet{kavouras2023fairness}) employ the term \textit{burden} to refer to the average cost itself. However, in an earlier work by \citet{milli2019social}, \textit{burden} is defined as the average cost incurred by instances that have a true positive class label. In this paper, we adopt the latter definition. }  
\textit{Social burden} measures the expected cost incurred by individuals who are already qualified (true positives) but receive an unfavorable prediction by a classifier $f$ \cite{milli2019social}.
The social burden is zero if the classifier positively classifies all qualified individuals without requiring changes to their non-sensitive attributes.
Conversely, the social burden is greater than zero when qualified individuals are assigned a negative classification and forced to alter their features to be recognized as valid.
Formally, we can define the social burden as:

\begin{definition}
    Let $f$ be a classifier and $g_f$ a recourse algorithm.
    Given a sensitive group $s\in\mathcal{S}$, let $\delta((x, S=s), g_f(x))$ be the cost of applying the recourse provided by $g_f$.
    Then, the expected social burden $B_{f,g}^{s}$ can be defined as:
    %
     \begin{equation}
         \underbrace{\mathbb{E}[\delta((X,s), g_f(X)) ]}_{\text{Exp. cost for inst. with  y=1, f(x)=0}}
        (1-\underbrace{P(f(X) = 1 \mid S = s, Y = 1)}_{\text{True Positive Rate (TPR)}})
        \label{eq:burden_eop}
    \end{equation}
    with the expectation over $P(X \mid S = s, Y = 1, f(X) = 0)$, and $P(f(X) = 1 \mid S = s, Y = 1)$ indicates the \textit{true positive rate} (TPR) for the given sensitive group. 
\end{definition}

Similarly to Eqn.~\ref{eq:cost_group}, the instances obtaining a positive prediction do not count towards the overall burden. 
Eqn.~\ref{eq:burden_eop} underscores the importance of considering the predictive performance of the base classifier when evaluating algorithmic recourse, a factor often overlooked in the recourse literature but central to algorithmic fairness; especially, since numerous studies have shown that machine learning classifiers tend to exhibit unequal error rates (\textit{e.g.,} TPR) across sensitive groups \cite{pessach2023algorithmic}. When error rates differ across groups, the burden of recourse is also likely to be unevenly distributed, potentially amplifying existing disparities. However, it is important to note that satisfying \textit{equality of opportunity} (\textit{i.e.,} parity in TPR) does not guarantee a small gap in social burden.

\subsection{Gap-Based Metrics can Mask Unfairness}

Unfairness in recourse can be quantified by measuring \textit{gaps} in cost or social burden across sensitive groups: $\Delta C_{f,g} = \max_{s,s' \in \mathcal{S}} |C^s_{f,g} - C^{s'}_{f,g}|$ or  $\Delta B_{f,g} = \max_{s,s' \in \mathcal{S}}  |B^s_{f,g} - B^{s'}_{f,g}|$. According to Equations~\ref{eq:cost_group} and \ref{eq:burden_eop}, these disparities stem from three main sources. For cost gaps, the causes include: (a) disparities in AR, as captured by statistical parity \cite{dwork2012fairness}; (b) differences in the cost function, which may arise from unequal opportunities or socio-economic conditions; and (c) variations in the distribution $P(X \mid S, f(X) = 0)$, reflecting potential population skews. For social burden gaps, they arise from: (a) disparities in true positive rates, captured by equality of opportunity \cite{hardt2016equality}, (b) unequal cost functions, and (c) skews in the distribution of qualified individuals denied recourse $P(X \mid S, Y = 1, f(X) = 0)$. Importantly, this illustrates that satisfying equality of opportunity (statistical parity) does not guarantee an equal distribution of the social burden (recourse costs) across groups.

Although comparing the \textit{gap of statistical measures} across groups is a common approach in the fairness literature, it can obscure meaningful underlying unfairness.
In fairness for prediction, a well-known limitation of parity-based definitions is that parity can be achieved by \textit{degrading} performance for the privileged group without improving outcomes for the unprivileged one \cite{martinez2020minimax, diana2021minimax}.
As a result, groups may exhibit similar expected values for a given metric, even when those values are far from optimal.
In recourse, this issue is especially critical when evaluating social burden, which ideally should be as low as possible (preferably zero).
A small gap in burden does not necessarily indicate fairness, as parity might be reached by \textit{increasing the burden for the privileged group} rather than reducing it for the unprivileged one.
However, in many real-world contexts, such as access to essential goods and services, sacrificing overall well-being is not acceptable.
Therefore, fairness efforts should aim to reduce the burden for all groups, rather than simply equalizing it.

We propose drawing inspiration from approaches in the fairness-in-prediction literature that address this limitation and complement difference-based disparity measures with a Rawlsian (\textit{i.e.,} minimax) perspective~\cite{binns2018fairness}.
For instance, in addition to evaluating the gap in social burden, we can examine the \textit{worst-group social burden} defined as $\max_{s \in \mathcal{S}} B_{f,g}^{s}$.
This perspective's benefits are twofold: (a) it helps identify the most vulnerable group, and (b) it serves as a base for the development of fairness-enhancing recourse frameworks that go beyond minimizing burden disparities, aiming instead to reduce the social burden for all groups.

\begin{algorithm}[t]
\caption{\texttt{MISOB}: MInimax SOcial Burden}
\begin{algorithmic}[1]
\REQUIRE $\mathcal{D} = \{(x^i, y^i)\}_{i=1}^N$, $T\in \mathbb{N}$, and $\alpha \in \mathbb{R}^+$
\STATE Pre-train $f^{(0)}$ using $\mathcal{D}$ \hfill //Warm-up Phase
\FOR{$t = 1$ to $T$}
    \STATE $\mathcal{Q} = \{\} $
    \FOR{$i=0$ to $N$}
        \STATE $b_{f,g}^i \gets \delta(x^i, g_f(x^i)) \mathds{1}\{ y^i=1\}$ //Compute burden
        \STATE $\mathcal{Q} \gets \mathcal{Q} \cup b_{f,g}^i$
    \ENDFOR
    \STATE $f^{(t)} \gets \arg\min_{f \in \mathcal{F}} \frac{1}{N} \sum_{i=1}^N \phi(i, \mathcal{Q}, \alpha) \cdot \ell(f(x^i), y^i)$
\ENDFOR
\RETURN $f^{(T)}$
\end{algorithmic}
\label{alg:misob}
\end{algorithm}

\begin{table*}[t]
\caption{ 
\textbf{Empirical average test results on the \textsc{Adult} dataset across different sensitive group characterizations.}
The best results for each setting are highlighted in bold, and we report whether the target metric should increase (${\uparrow}$) or decrease (${\downarrow}$).
For \texttt{MISOB}, we set $\alpha=0.3$ in all experiments (\textit{cf.} Eqn.~\ref{eq:weighting}).
Each cell shows the average and standard deviation over 10 runs. 
} 
\label{tab:results_adult}
\centering
\begin{sc}
\resizebox{\textwidth}{!}{%
\begin{tabular}{cccccccccccc}
\toprule
\multicolumn{1}{c}{Sensitive}    & Recourse & Fairness              & acc                        & \multicolumn{2}{c}{Burden (Eq.~\ref{eq:burden_eop})}   & \multicolumn{2}{c}{TPR}                                 & \multicolumn{2}{c}{Cost (Eq.~\ref{eq:cost_group})}         & \multicolumn{2}{c}{AR}                                  \\
\multicolumn{1}{c}{Attribute(s)}    & Method   & strategy              & overall ${\uparrow}$       & worst ${\downarrow}$       & $\Delta$ ${|\downarrow|}$  & worst ${\uparrow}$         & $\Delta$ ${|\downarrow|}$  & worst ${\downarrow}$         & $\Delta$ ${|\downarrow|}$    & worst ${\uparrow}$         & $\Delta$ ${|\downarrow|}$  \\ \midrule
\multirow{9}{*}{Race}            & GS       & -                     & 0.81$_{\pm 0.02}$          & 4.56$_{\pm 0.01}$          & \textbf{0.03}$_{\pm 0.02}$ & 0.27$_{\pm 0.03}$          & 0.08$_{\pm 0.08}$          & 115.69$_{\pm 1.19}$          & 28.37$_{\pm 1.95}$           & 0.06$_{\pm 0.01}$          & 0.06$_{\pm 0.02}$          \\
                                 & GS       & \texttt{POSTPRO}        & 0.80$_{\pm 0.01}$          & 4.96$_{\pm 3.49}$          & 0.61$_{\pm 2.78}$          & 0.37$_{\pm 0.51}$          & \textbf{0.00}$_{\pm 0.52}$ & 98.40$_{\pm 56.73}$          & \textbf{17.19}$_{\pm 42.75}$ & \textbf{0.37}$_{\pm 0.53}$ & \textbf{0.01}$_{\pm 0.54}$ \\
                                 & GS       & \texttt{MISOB}         & \textbf{0.82}$_{\pm 0.01}$ & \textbf{3.01}$_{\pm 0.32}$ & 0.85$_{\pm 0.46}$          & \textbf{0.52}$_{\pm 0.02}$ & 0.11$_{\pm 0.04}$          & \textbf{93.06}$_{\pm 0.98}$  & 27.38$_{\pm 1.63}$           & 0.15$_{\pm 0.01}$          & 0.12$_{\pm 0.02}$          \\ \cmidrule(l){2-12} 
                                 & WT       & -                     & 0.81$_{\pm 0.02}$          & 1.28$_{\pm 0.14}$          & \textbf{0.01}$_{\pm 0.18}$ & 0.27$_{\pm 0.03}$          & 0.08$_{\pm 0.08}$          & 38.27$_{\pm 0.57}$           & 12.32$_{\pm 0.72}$           & 0.06$_{\pm 0.01}$          & 0.06$_{\pm 0.02}$          \\
                                 & WT       & \texttt{POSTPRO}        & 0.80$_{\pm 0.01}$          & 1.55$_{\pm 0.27}$          & \textbf{0.01}$_{\pm 0.47}$ & 0.37$_{\pm 0.51}$          & \textbf{0.00}$_{\pm 0.52}$ & 39.71$_{\pm 5.03}$           & 13.22$_{\pm 2.10}$           & \textbf{0.37}$_{\pm 0.53}$ & \textbf{0.01}$_{\pm 0.54}$ \\
                                 & WT       & \texttt{MISOB}          & \textbf{0.82}$_{\pm 0.01}$ & \textbf{0.79}$_{\pm 0.00}$ & 0.16$_{\pm 0.00}$          & \textbf{0.59}$_{\pm 0.04}$ & 0.02$_{\pm 0.06}$          & \textbf{30.77}$_{\pm 0.23}$  & \textbf{11.66}$_{\pm 0.45}$  & 0.16$_{\pm 0.02}$          & 0.08$_{\pm 0.04}$          \\ \cmidrule(l){2-12} 
                                 & CCHVAE   & -                     & \textbf{0.81}$_{\pm 0.02}$ & 6.25$_{\pm 0.46}$          & \textbf{0.16}$_{\pm 0.89}$ & 0.27$_{\pm 0.03}$          & 0.08$_{\pm 0.08}$          & 119.99$_{\pm 3.78}$          & \textbf{27.01}$_{\pm 7.52}$  & 0.06$_{\pm 0.01}$          & 0.06$_{\pm 0.02}$          \\
                                 & CCHVAE   & \texttt{POSTPRO}        & 0.80$_{\pm 0.01}$          & 11.20$_{\pm 1.89}$         & 3.06$_{\pm 2.96}$          & 0.37$_{\pm 0.51}$          & \textbf{0.00}$_{\pm 0.52}$ & 120.01$_{\pm 8.94}$          & 21.22$_{\pm 9.38}$           & \textbf{0.37}$_{\pm 0.53}$ & \textbf{0.01}$_{\pm 0.54}$ \\
                                 & CCHVAE   & \texttt{MISOB}          & \textbf{0.81}$_{\pm 0.01}$ & \textbf{4.03}$_{\pm 0.44}$ & 0.42$_{\pm 0.54}$          & \textbf{0.48}$_{\pm 0.03}$ & 0.19$_{\pm 0.05}$          & \textbf{105.10}$_{\pm 1.14}$ & 27.63$_{\pm 1.98}$           & 0.12$_{\pm 0.01}$          & 0.16$_{\pm 0.02}$          \\ \midrule
\multirow{9}{*}{Gender}          & GS       & -                     & 0.81$_{\pm 0.02}$          & 4.73$_{\pm 0.39}$          & 0.54$_{\pm 0.58}$          & 0.32$_{\pm 0.05}$          & 0.08$_{\pm 0.04}$          & 109.75$_{\pm 2.96}$          & 28.76$_{\pm 3.72}$           & 0.06$_{\pm 0.03}$          & 0.07$_{\pm 0.04}$          \\
                                 & GS       & \texttt{POSTPRO}        & 0.80$_{\pm 0.01}$          & 24.46$_{\pm 25.08}$        & 13.98$_{\pm 8.03}$         & 0.00$_{\pm 0.00}$          & \textbf{0.00}$_{\pm 0.00}$ & 145.47$_{\pm 47.36}$         & \textbf{28.17}$_{\pm 63.66}$ & 0.00$_{\pm 0.00}$          & \textbf{0.00}$_{\pm 0.00}$ \\
                                 & GS       & \texttt{MISOB}          & \textbf{0.82}$_{\pm 0.01}$ & \textbf{2.59}$_{\pm 0.00}$ & \textbf{0.33}$_{\pm 0.00}$ & \textbf{0.55}$_{\pm 0.01}$ & 0.04$_{\pm 0.02}$          & \textbf{94.26}$_{\pm 0.71}$  & 32.49$_{\pm 1.13}$           & \textbf{0.10}$_{\pm 0.01}$ & 0.16$_{\pm 0.01}$          \\ \cmidrule(l){2-12} 
                                 & WT       & -                     & 0.81$_{\pm 0.02}$          & 1.29$_{\pm 0.03}$          & \textbf{0.04}$_{\pm 0.05}$ & 0.32$_{\pm 0.05}$          & 0.08$_{\pm 0.08}$          & 43.06$_{\pm 0.37}$           & 22.88$_{\pm 0.55}$           & 0.06$_{\pm 0.03}$          & 0.07$_{\pm 0.04}$          \\
                                 & WT       & \texttt{POSTPRO}        & 0.80$_{\pm 0.01}$          & 1.65$_{\pm 0.39}$          & 0.33$_{\pm 0.22}$          & 0.00$_{\pm 0.00}$          & \textbf{0.00}$_{\pm 0.00}$ & 42.40$_{\pm 3.59}$           & 20.99$_{\pm 2.12}$           & 0.00$_{\pm 0.00}$          & \textbf{0.00}$_{\pm 0.00}$ \\
                                 & WT       & \texttt{MISOB}          & \textbf{0.82}$_{\pm 0.01}$ & \textbf{0.79}$_{\pm 0.06}$ & 0.21$_{\pm 0.08}$          & \textbf{0.55}$_{\pm 0.03}$ & 0.04$_{\pm 0.04}$          & \textbf{34.36}$_{\pm 0.31}$  & \textbf{20.29}$_{\pm 0.50}$  & \textbf{0.10}$_{\pm 0.03}$ & 0.17$_{\pm 0.04}$          \\ \cmidrule(l){2-12} 
                                 & CCHVAE   & -                     & \textbf{0.81}$_{\pm 0.02}$ & 6.43$_{\pm 0.64}$          & \textbf{0.57}$_{\pm 1.02}$ & 0.32$_{\pm 0.05}$          & 0.08$_{\pm 0.08}$          & 116.69$_{\pm 4.83}$          & 31.12$_{\pm 6.87}$           & 0.06$_{\pm 0.03}$          & 0.07$_{\pm 0.04}$          \\
                                 & CCHVAE   & \texttt{POSTPRO}        & 0.80$_{\pm 0.01}$          & 12.97$_{\pm 4.20}$         & 6.61$_{\pm 4.78}$          & 0.00$_{\pm 0.00}$          & \textbf{0.00}$_{\pm 0.00}$ & 119.88$_{\pm 7.24}$          & \textbf{27.32}$_{\pm 12.12}$ & 0.00$_{\pm 0.00}$          & \textbf{0.00}$_{\pm 0.00}$ \\
                                 & CCHVAE   & \texttt{MISOB}          & \textbf{0.81}$_{\pm 0.01}$ & \textbf{4.18}$_{\pm 0.27}$ & 0.77$_{\pm 0.43}$          & \textbf{0.55}$_{\pm 0.03}$ & 0.12$_{\pm 0.05}$          & \textbf{103.98}$_{\pm 1.52}$ & 35.58$_{\pm 2.69}$           & \textbf{0.11}$_{\pm 0.02}$ & 0.22$_{\pm 0.03}$          \\ \midrule
\multirow{9}{*}{\parbox{1.3cm}{Race \& Gender}} & GS       & -                     & 0.81$_{\pm 0.02}$          & 5.13$_{\pm 0.66}$          & 1.79$_{\pm 1.22}$          & 0.20$_{\pm 0.04}$          & 0.22$_{\pm 0.07}$          & 124.86$_{\pm 2.33}$          & \textbf{48.63}$_{\pm 3.52}$  & 0.02$_{\pm 0.01}$          & 0.11$_{\pm 0.02}$          \\
                                 & GS       & \texttt{POSTPRO}        & 0.80$_{\pm 0.01}$          & 5.12$_{\pm 2.26}$          & 3.13$_{\pm 2.81}$          & 0.00$_{\pm 0.00}$          & \textbf{0.00}$_{\pm 0.00}$ & 128.05$_{\pm 64.21}$         & 91.55$_{\pm 42.15}$          & 0.00$_{\pm 0.00}$          & \textbf{0.00}$_{\pm 0.00}$ \\
                                 & GS       & \texttt{MISOB}          & \textbf{0.82}$_{\pm 0.01}$ & \textbf{3.37}$_{\pm 0.70}$ & \textbf{1.32}$_{\pm 1.00}$ & \textbf{0.46}$_{\pm 0.05}$ & \textbf{0.19}$_{\pm 0.07}$ & \textbf{109.95}$_{\pm 2.35}$ & 51.28$_{\pm 3.33}$           & \textbf{0.06}$_{\pm 0.01}$ & 0.23$_{\pm 0.02}$          \\ \cmidrule(l){2-12} 
                                 & WT       & -                     & 0.81$_{\pm 0.02}$          & 1.40$_{\pm 0.36}$          & \textbf{0.27}$_{\pm 0.54}$ & 0.20$_{\pm 0.04}$          & 0.22$_{\pm 0.07}$          & 49.71$_{\pm 1.38}$           & 30.81$_{\pm 2.40}$           & 0.02$_{\pm 0.01}$          & 0.11$_{\pm 0.02}$          \\
                                 & WT       & \texttt{POSTPRO}        & 0.80$_{\pm 0.01}$          & 1.94$_{\pm 0.75}$          & 0.90$_{\pm 0.48}$          & 0.00$_{\pm 0.00}$          & \textbf{0.00}$_{\pm 0.00}$ & 51.48$_{\pm 8.83}$           & 30.81$_{\pm 4.33}$           & 0.00$_{\pm 0.00}$          & \textbf{0.00}$_{\pm 0.00}$ \\
                                 & WT       & \texttt{MISOB}          & \textbf{0.82}$_{\pm 0.01}$ & \textbf{0.98}$_{\pm 0.08}$ & 0.45$_{\pm 0.12}$          & \textbf{0.34}$_{\pm 0.05}$ & 0.25$_{\pm 0.08}$          & \textbf{41.43}$_{\pm 0.66}$  & \textbf{28.59}$_{\pm 0.93}$  & \textbf{0.04}$_{\pm 0.03}$ & 0.25$_{\pm 0.04}$          \\ \cmidrule(l){2-12} 
                                 & CCHVAE   & -                     & \textbf{0.81}$_{\pm 0.02}$ & 7.37$_{\pm 0.51}$          & 3.53$_{\pm 1.01}$          & 0.20$_{\pm 0.04}$          & 0.22$_{\pm 0.07}$          & 133.32$_{\pm 1.58}$          & 53.24$_{\pm 3.16}$           & 0.02$_{\pm 0.01}$          & 0.11$_{\pm 0.02}$          \\
                                 & CCHVAE   & \texttt{POSTPRO} & 0.80$_{\pm 0.01}$          & 18.25$_{\pm 1.47}$         & 13.07$_{\pm 3.64}$         & 0.00$_{\pm 0.00}$          & \textbf{0.00}$_{\pm 0.00}$ & 149.20$_{\pm 10.24}$         & 66.95$_{\pm 10.89}$          & 0.00$_{\pm 0.00}$          & \textbf{0.00}$_{\pm 0.00}$ \\
                                 & CCHVAE   & \texttt{MISOB}          & \textbf{0.81}$_{\pm 0.01}$ & \textbf{5.36}$_{\pm 0.57}$ & \textbf{2.11}$_{\pm 1.09}$ & \textbf{0.35}$_{\pm 0.05}$ & 0.33$_{\pm 0.08}$          & \textbf{116.93}$_{\pm 0.92}$ & \textbf{52.78}$_{\pm 1.67}$  & \textbf{0.04}$_{\pm 0.02}$ & 0.31$_{\pm 0.03}$          \\ \bottomrule
\end{tabular}%
}
\end{sc}
\end{table*}

\begin{figure*}[t] 
    \centering
    \begin{subfigure}[b]{\textwidth} 
        \centering
        \includegraphics[width=0.8\columnwidth]{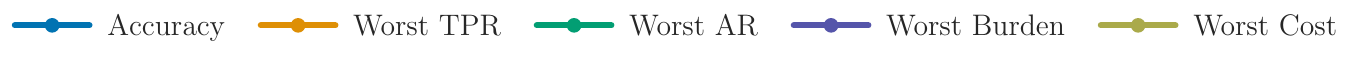}
    \end{subfigure}    
    \begin{subfigure}[b]{0.24\textwidth} 
        \centering
        \includegraphics[width=\columnwidth]{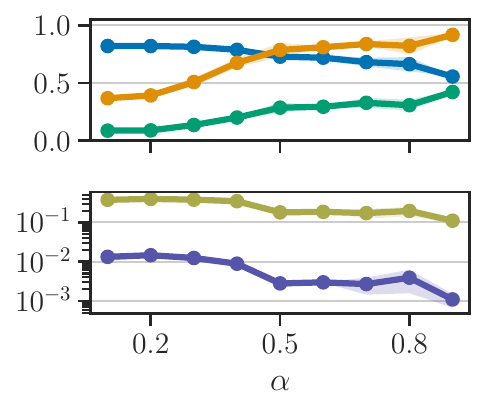}
        \caption{WT - Race}
    \end{subfigure}
    \hfill
    \begin{subfigure}[b]{0.24\textwidth} 
        \centering
        \includegraphics[width=\columnwidth]{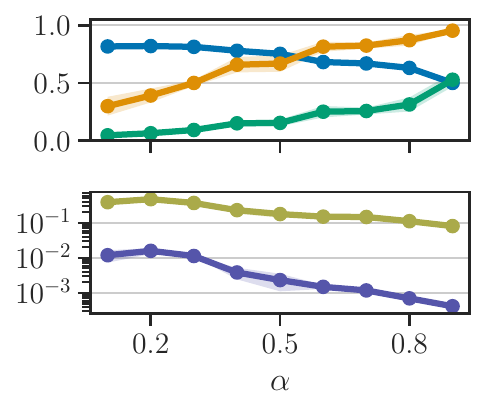}
        \caption{WT - Gender}
    \end{subfigure}
    \hfill
    \begin{subfigure}[b]{0.24\textwidth} 
        \centering
        \includegraphics[width=\columnwidth]{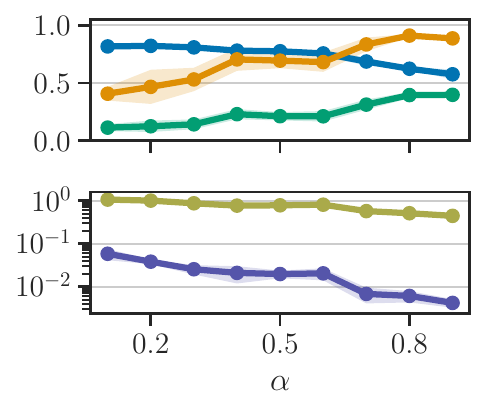}
        \caption{GS - Race}
    \end{subfigure}
    \hfill
    \begin{subfigure}[b]{0.24\textwidth} 
        \centering
        \includegraphics[width=\columnwidth]{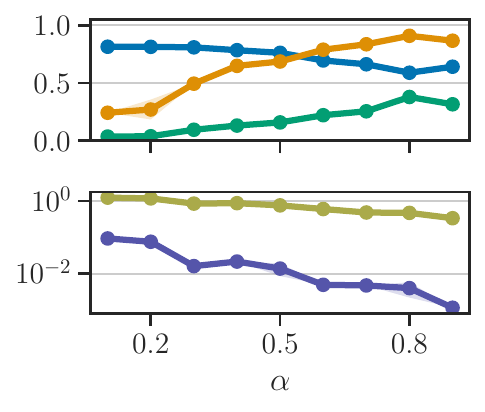}
        \caption{GS - Gender}
    \end{subfigure}
    \caption{\textbf{Empirical evaluation of \texttt{MISOB} for different} $\alpha \in \{0.1, \ldots, 1.0\}$\textbf{.}
    We report the average results and standard deviation (shaded area) over 10 runs, with WT \cite{wachter2017counterfactual} and GS \cite{laugel2017inverse}, on the \textsc{Adult} dataset with \textit{race} and \textit{gender} as the sensitive attributes.
    The worst burden and cost (Eqn.~\ref{eqn:eval_metrics}) are in log scale.
    In brief, increasing $\alpha$, which favors high-burden instances, improves fairness guarantees, but large $\alpha$ values may hurt overall accuracy.
    }
    \label{fig:GC_ablation_C}
\end{figure*}
\section{\texttt{MISOB}: MInimax SOcial Burden}
\label{sec:miso}

In this section, we introduce a simple yet effective iterative training procedure, \texttt{MISOB} (MInimax SOcial Burden), that empirically reduces social burden (Eqn.~\ref{eq:burden_eop}) without degrading predictive performance and \textit{without requiring access to sensitive group attributes during training or inference}.
The method is outlined in Algorithm~\ref{alg:misob}.
Given a pre-trained classifier (line 1)\footnote{Empirically, we observed that pretraining a base classifier to achieve reasonable accuracy before applying \texttt{MISOB} contributes to a more stable optimization process.}, the key idea is to incorporate social burden estimates directly into the training loop.
At each iteration, we compute the social burden for each instance (line 5), and we save the result within $\mathcal{Q}$ (line 6).
The model is then updated by minimizing a \textit{(weighted) loss function} $\ell(f(x^i), y^i)$ (line 8), where weights are determined by the estimated burden $\phi(i, \mathcal{Q})$.
Specifically, we propose assigning weights to instances $x^i$ based on the following scheme: 
\begin{equation}
\phi(i, \mathcal{Q}, \alpha) = 1 + \alpha N \dfrac{b_{f,g_f}^i}{\sum_{j \in |\mathcal{Q}|} b_{f,g_f}^j} \mathds{1}\{\beta>0\}
\label{eq:weighting}
\end{equation}
where $N$ is the dataset (or batch) size, $\beta=\sum_{j \in |\mathcal{Q}|} b_{f,g_f}^j$ is the total burden with $b_{f,g}^i = \delta(x^i, g_f(x^i)) \mathds{1}\{ y^i=1\} $ being the burden of instance $i$, and $\alpha$ is a hyperparameter controlling the influence of the burden term.
Under this weighting scheme, each instance $x^i$ is assigned a weight proportional to its contribution to the total burden across the dataset (or batch).
In the degenerate case, where the total burden is zero, all instances are assigned equal weights.
This reweighting scheme amplifies the influence of \textit{high-burden instances} during training, encouraging the classifier to prioritize decisions that reduce unnecessary or costly adjustments imposed on individuals under the current decision rule.

The proposed framework offers several advantages that enhance its practicality and broad applicability.\footnote{The runtime of Algorithm~\ref{alg:misob} is determined by three factors: (i) the number of retraining steps $T$, (ii) the dataset size $N$, and (iii) the run-time complexity of the recourse method $O(K)$. 
In practice, datasets are typically much larger than the number of retraining iterations, and if a simple recourse method is used, we generally have $K << T << N$. 
As a result, the overall computational complexity is $O(N^3)$.
Efficiency can be further improved through batching, parallelization, or approximating recourse computations.
}
First, the method is \textit{agnostic} to both the base classifier and the chosen recourse strategy, allowing seamless integration with a wide range of existing approaches. It also operates \textit{without access to sensitive attributes during training}, avoiding legal, ethical, and practical challenges associated with collecting or using such data.
Further, \texttt{MISOB} not only improves deployability but also naturally addresses intersectional fairness.
Because group definitions are not required a-priori, intersectional subgroups can be introduced post-hoc at \textit{evaluation time}, enabling fairness assessments across multiple dimensions \textit{without retraining}.
This flexibility makes the framework well-suited for real-world applications where fairness concerns span diverse and evolving group identities.

\section{Empirical Evaluation}
\label{sec:experiments}

In this section, we evaluate the effectiveness of \texttt{MISOB} in reducing the social burden across sensitive groups.
Refer to Appendix~\ref{ap:experimental_setting} for further details on the experimental setting. 

\paragraph{Experimental Setting.}
We consider the real-world dataset \textsc{Adult}~\cite{adult_2} with sensitive groups defined by different characterizations of the sensitive attributes (\textit{race} and \textit{gender}), considering also intersectional groups. 
For the model, we consider a simple feed-forward neural network. 
We have chosen this base classifier because it supports multiple different recourse methods proposed in the literature.
Further experiments with a linear regression model and other datasets can be found in Appendix \ref{app:additional_results}.
In our evaluation, we consider \texttt{MISOB} and \texttt{POSTPRO}~\citep{hardt2016equality}, which enhance prediction fairness in terms of \textit{equality of opportunity}. 
We employ \texttt{POSTPRO} because it is compatible with any base classifier, enabling fair comparisons with \texttt{MISOB} under identical predictive models.
We consider three widely adopted recourse methods, selected for their benchmarking popularity, available implementations, and ease of integration in an agnostic setting: Growing Spheres (GS) \cite{laugel2017inverse}, the method by \citet{wachter2017counterfactual} (WT), and the Counterfactual Conditional Heterogeneous Variational Autoencoder (CCHVAE) by \citet{pawelczyk2020learning}. 
We use the implementations and hyperparameters from the benchmark by \citet{pawelczyk1carla}, and we repeat the experiments over 10 random train/test splits.
We consider the worst-group values for various metrics:
\begin{gather}
    \min_{s \in \mathcal{S}} P(f(X=x)=1 \mid S=s, Y=1) \; \text{(Worst TPR)} \\
    \min_{s \in \mathcal{S}} P(f(X=x)=1 \mid S=s) \; \text{(Worst AR)} \\
    \min_{s \in \mathcal{S}} C_{f,g}^s \; \text{(Worst Cost)} \qquad \min_{s \in \mathcal{S}} B_{f,g}^s \; \text{(Worst Burden)}
    \label{eqn:eval_metrics}
\end{gather}
where the recourse cost $\delta$ is quantified using the $\ell_2$ distance, and their maximum disparity (denoted by $\Delta$) across groups:
\begin{gather}
    \max_{s,s' \in \mathcal{S}} |C_{f,g}^s - C_{f,g}^{s'}| \qquad \max_{s,s' \in \mathcal{S}} |B_{f,g}^s - B_{f,g}^{s'}| \\
    \max_{s,s' \in \mathcal{S}} |\text{TPR}^s - \text{TPR}^{s'}| \qquad \max_{s,s' \in \mathcal{S}} |\text{AR}^s - \text{AR}^{s'}|
\end{gather}

Importantly, our method operates \textit{entirely without access to sensitive attributes}.
Note that computing fairness metrics-whether for prediction or recourse-requires access to sensitive attributes, which is an inherent limitation of these metrics themselves \cite{buyl2024inherent}.

\paragraph{Results.}
Table~\ref{tab:results_adult} shows that training the base classifier under the \texttt{MISOB} framework improves both predictive and recourse-related performance metrics across all groups, without compromising overall accuracy.
Notably, \texttt{MISOB} enhances \textit{worst-case group outcomes} and, in many cases, reduces disparities across groups, achieving fairness improvements without degrading the performance for privileged groups.
These results also highlight an important limitation of gap-based fairness metrics: small gaps can mask universally poor outcomes across groups, giving a misleading sense of fairness.
In contrast, \texttt{MISOB} can produce configurations with slightly larger gaps but better metric values for every group, thereby promoting greater overall well-being.
Moreover, the results show that improving fairness at the prediction level does not guarantee fairer recourse outcomes and can even exacerbate them.
In particular, enforcing parity in TPR through \texttt{POSTPRO} significantly increases the burden and cost for decision subjects.
This occurs when prediction-level fairness improvements are misaligned with fairness improvements at recourse.
On the contrary, the holistic perspective of \texttt{MISOB} enables effective improvements in group TPR, while reducing the cost and burden of individuals, underscoring the importance of considering the entire pipeline.
Besides, note that \texttt{POSTPRO} requires training a \textit{separate model} for each sensitive group characterization, whereas \texttt{MISOB}, being blind to sensitive information, needs to be trained only once and can be evaluated across multiple group characterizations.

The performance and fairness of \texttt{MISOB} are governed by the emphasis placed on high-burden individuals, controlled by the weighting parameter $\alpha$ from Eqn.~\ref{eq:weighting}.
Fig.~\ref{fig:GC_ablation_C} shows that increasing $\alpha$ (\textit{i.e.,} the importance of high-burden instances) strengthens minimax fairness across both prediction and recourse by improving worst-group metrics. 
However, larger values of $\alpha$ ($\alpha > 0.5$) can reduce overall accuracy due to the overemphasis on high-burden cases.
In practice, $\alpha$ governs the trade-off between fairness and accuracy, allowing users to select the value that best aligns with their objectives.

\section{Discussion and Limitations}

%
Our work introduces a novel framework to compute fairness metrics in a recourse-aware setting and empirically validates the effectiveness of \texttt{MISOB}.
Analyzing the convergence properties of the training procedure (\textit{e.g.,} identifying stationary points) would be a valuable direction for further research.
One approach might be to frame recourse fairness as a two-player Stackelberg game, following the \textit{strategic classification} perspective \cite{chen2023learning}.
Additionally, applying calibration techniques \cite{guo2017calibration} could improve prediction reliability (\textit{cf.} Eqn.~\ref{eq:burden_eop}), which in turn may further enhance fairness when combined with \texttt{MISOB}.
Finally, further fairness concerns may emerge due to model updates, dataset shifts~\cite{castelnovo2021towards}, prediction performativity \cite{liu2018delayed, perdomo2020performative}, or recourse recommendations that are not robust over time~\cite{detoni2025temporal}.
Extending our approach to measure and optimize the social burden of recourse in dynamic settings is, therefore, a natural path for future research.
Lastly, we aim to explore more realistic cost models that better reflect the real world's complexities and assess how these influence fairness guarantees~\cite{tominaga2024reassessing,esfahani2024preference}.

\section{Conclusions}
\label{sec:conclusions}

In this work, we study the unfairness that can arise in automated decision-making pipelines that provide recourse. We provide a theoretical characterization of the sources of unfairness in algorithmic recourse, \textit{formally linking fairness guarantees in recourse and prediction}, and further highlight the limitations of the standard equal cost paradigm.
To address these challenges, we introduce \texttt{MISOB}, a novel framework grounded in the notion of \textit{social burden}, which enables a holistic treatment of fairness and goes beyond a gap-based characterization of fairness guarantees.
\texttt{MISOB} is broadly applicable: it is compatible with many classifier–recourse pairs and does not require access to sensitive attributes during training or inference.
Empirical results show that improving fairness solely at the prediction level is often insufficient and can even considerably increase the cost and burden of decision subjects.
In contrast, our approach leads to consistent joint fairness improvements in prediction and recourse, without compromising overall performance.

\section*{Acknowledgments}

We would like to thank all the reviewers for their insightful feedback.
We also thank Mikel Malagon for the technical support and for reading preliminary versions of the paper.
This research was funded by the European Union. Views and opinions expressed are however those of the author(s) only and do not necessarily reflect those of the European Union or the European Health and Digital Executive Agency (HaDEA). Neither the European Union nor the granting authority can be held responsible for them. This work is supported by the European Research Council under the European Union's Horizon 2020 research and innovation programme Grant Agreement no. 851538 - BayesianGDPR, Horizon Europe research and innovation programme Grant Agreement no. 101120763 - TANGO, and no. 101120237 - ELIAS. This work is also supported by the Basque Government under grant IT1504-22 and through the BERC 2022-2025 program; by the Spanish Ministry of Science and Innovation under the grant PID2022-137442NB-I00, and through
BCAM Severo Ochoa accreditation CEX2021-001142-S / MICIN / AEI / 10.13039/501100011033.
Lastly, this work was also supported by Ministero delle Imprese e del Made in Italy (IPCEI Cloud DM 27 giugno 2022 – IPCEI-CL-0000007), funded by the European Commission under the NextGeneration EU programme.

\bibliography{aaai2026}

\onecolumn

\renewcommand{\thesection}{\Alph{section}}

\appendix

\section*{\huge Appendix}
\vspace{2em}

\section{Proofs}
\label{app:proofs}

\subsection{Proof of Equation~\ref{eq:cost_group}}
\label{app:sec:expected_cost}

Let us denote the recourse cost with the random variable $Z = \delta((X,S), g_f(X))$.
Further, we denote with $P(f(X)=0 \mid S=s)$ the likelihood of being classified negatively, conditioned on the protected group $s \in \mathcal{S}$. 
Then, the expected recourse cost for a given protected group $s \in \mathcal{S}$ can be defined as follows:
\begin{align}
    \mathbb{E}[Z \mid S=s] &= \sum_{x \in \mathcal{X}} Z(x, s) P(X=x \mid S=s) \\
    &= \sum_{x' \in \mathcal{X} : f(x')=0} Z(x', s) P(X=x' \mid S=s) + \sum_{x'' \in \mathcal{X} : f(x'')=1} Z(x'', s) P(X=x'' \mid S=s) \\
    &\overset{(i)}{=} \sum_{x' \in \mathcal{X} : f(x')=0} Z(x', s) P(X=x' \mid S=s) \\
    &= \frac{P(f(X)=0 \mid S=s)}{P(f(X)=0 \mid S=s)}\sum_{x' \in \mathcal{X} : f(x')=0} Z(x', s) P(X=x' \mid S=s) \\
    &= P(f(X)=0 \mid S=s)\sum_{x' \in \mathcal{X} : f(x')=0} Z(x', s) \frac{P(X=x' \mid S=s)}{P(f(X)=0 \mid S=s)} \\
    &= P(f(X)=0 \mid S=s)\sum_{x' \in \mathcal{X} : f(x')=0} Z(x', s) \frac{P(X=x' \mid S=s)}{\sum_{\mathcal{X}} \mathds{1}\{f(x''')=0\}P(X=x''' \mid S=s)} \\
    &= P(f(X)=0 \mid S=s)\sum_{x' \in \mathcal{X} : f(x')=0} Z(x', s) \frac{P(X=x' \mid S=s)}{\sum_{x''' \in \mathcal{X} : f(x''')=0}P(X=x''' \mid S=s)} \\
    &= P(f(X)=0 \mid S=s)\sum_{x' \in \mathcal{X} : f(x')=0} Z(x', s) P(X=x' \mid S=s, f(X=x')=0) \\
    &= P(f(X)=0 \mid S=s)\mathbb{E}[Z \mid S=s, f(X)=0] \\
    &= \mathbb{E}[Z \mid S=s, f(X)=0] (1-P(f(X)=1 \mid S=s)
\end{align}
where $(i)$ arises from assuming that those instances that have received a positive classification (\textit{i.e.,} $x' \in \mathcal{X} : f(x')=1$) do not change their features and, therefore, $\delta((x,s),g_f(x))=0$.
Then, by substituting $Z$, we arrive at the expected recourse cost: 
\begin{align}
    C_{f,g}^{s} &= \mathbb{E}[Z \mid S=s, f(X)=0] (1-P(f(X)=1 \mid S=s) \\
    &= \mathbb{E}[\delta((X,S=s),g_f(X)) \mid S=s, f(X)=0] (1-P(f(X)=1 \mid S=s))
\end{align}
In summary, the actual expected cost of algorithmic recourse depends on two factors: (i) the effort required of individuals in the sensitive group $s\in\mathcal{S}$ who receive a negative classification, quantified by $\mathbb{E}[\delta((X,S=s),g_f(X)) \mid S=s, f(X)=0]$, and (ii) the probability that these individuals receive a negative outcome, $1-P(f(X)=1 \mid S=s)$.

\subsection{Proof of Equation~\ref{eq:burden_eop}}

We follow a similar setup to the one in Section~\ref{app:sec:expected_cost}.
Let us denote the recourse cost with the random variable $Z = \delta((X,S), g_f(X))$.
We denote with $P(f(X)=0 \mid S=s, Y=1)$ the likelihood of being incorrectly classified negatively by the model, conditioned on the protected group $s \in \mathcal{S}$, given that the ground truth outcome is positive $Y=1$.
Then, the \textit{expected social burden} for a given protected group $s \in \mathcal{S}$ can be defined similarly as in Section~\ref{app:sec:expected_cost}.
We report the full proof here for completeness:
\begingroup
\allowdisplaybreaks
\begin{align}
    \mathbb{E}[Z \mid S=s, Y=1] &= \sum_{x \in \mathcal{X}} Z(x, s) P(X=x \mid S=s, , Y=1) \\
    &= \sum_{x' \in \mathcal{X} : f(x')=0} Z(x', s) P(X=x' \mid S=s, Y=1) + \sum_{x'' \in \mathcal{X} : f(x'')=1} Z(x'', s) P(X=x'' \mid S=s, Y=1) \\
    &\overset{(i)}{=} \sum_{x' \in \mathcal{X} : f(x')=0} Z(x', s) P(X=x' \mid S=s, Y=1) \\
    &= \frac{P(f(X)=0 \mid S=s, Y=1)}{P(f(X)=0 \mid S=s, Y=1)}\sum_{x' \in \mathcal{X} : f(x')=0} Z(x', s) P(X=x' \mid S=s, Y=1) \\
    &= P(f(X)=0 \mid S=s, Y=1)\sum_{x' \in \mathcal{X} : f(x')=0} Z(x', s) \frac{P(X=x' \mid S=s, Y=1)}{P(f(X)=0 \mid S=s, Y=1)} \\
    &= P(f(X)=0 \mid S=s, Y=1)\sum_{x' \in \mathcal{X} : f(x')=0} Z(x', s) \frac{P(X=x' \mid S=s, Y=1)}{\sum_{\mathcal{X}} \mathds{1}\{f(x''')=0\}P(X=x''' \mid S=s, Y=1)} \\
    &= P(f(X)=0 \mid S=s, Y=1)\sum_{x' \in \mathcal{X} : f(x')=0} Z(x', s) \frac{P(X=x' \mid S=s, Y=1)}{\sum_{x''' \in \mathcal{X} : f(x''')=0}P(X=x''' \mid S=s, Y=1)} \\
    &= P(f(X)=0 \mid S=s, Y=1)\sum_{x' \in \mathcal{X} : f(x')=0} Z(x', s) P(X=x' \mid S=s, f(X=x')=0, Y=1) \\
    &= P(f(X)=0 \mid S=s, Y=1)\mathbb{E}[Z \mid S=s, f(X)=0, Y=1] \\
    &= \mathbb{E}[Z \mid S=s, f(X)=0, Y=1] (1-P(f(X)=1 \mid S=s, Y=1))
\end{align}
\endgroup
where $(i)$ arises from assuming that those instances that have received a positive classification (\textit{i.e.,} $x' \in \mathcal{X} : f(x')=1$) do not change their features and, therefore, $\delta((x,s),g_f(x))=0$.
Then, by substituting $Z$, we arrive at the expected social burden: 
\begin{align}
    B_{f,g}^{s} &= \mathbb{E}[Z \mid S=s, f(X)=0, Y=1] (1-P(f(X)=1 \mid S=s, Y=1)) \\
    &= \mathbb{E}[\delta((X,S=s),g_f(X)) \mid S=s, f(X)=0, Y=1] (1-P(f(X)=1 \mid S=s, Y=1))
\end{align}


\section{Further Implementation Details for the Experiments}
\label{ap:experimental_setting}

\subsection{Datasets}

In the paper, we used three fairly common datasets both in the recourse and fairness literature.
Specifically, we used \textsc{Adult} \cite{adult_2} for the experiments in Section~\ref{sec:experiments}, and \textsc{GiveMeSomeCredit}\footnote{\url{https://www.kaggle.com/c/GiveMeSomeCredit/}} and \textsc{German Credit} \cite{germancredit} for the additional experiments in Appendix~\ref{app:additional_results}.

\paragraph{Adult.} It is based on the data from the 1994 US Census, where the main goal is to predict whether an individual earns more than 50,000\$ per year.
The 14 features used to describe the instances include occupation, marital status, and education.
Furthermore, it contains sensitive information such as \textit{age}, \textit{gender}, and \textit{race}.
In our experiments, we consider sensitive groups based on gender and race, as well as \textit{intersectional groups} defined by their combination.
This dataset is publicly available in the UCI repository.\footnote{\url{http://archive.ics.uci.edu/ml/index.php}}
Although the dataset is originally divided into predefined training and test sets, we merge them and generate new train/test splits for our empirical evaluation.
We pre-process the dataset as detailed in \citet{pawelczyk1carla}.

\paragraph{Give Me Some Credit.}
It includes financial and demographic information for 150,000 individuals and is used to predict the likelihood of experiencing financial distress within two years.
Each record consists of 10 feature variables, such as age, monthly income, and credit indicators, and a binary target variable indicating whether the individual experienced a serious delinquency (90+ days past due) within that timeframe (0 = Yes, 1 = No).
For our fairness-aware analysis, we used \textit{age} as the sensitive attribute, dividing individuals into two groups: those aged 30 or younger, and those older than 30.
This distinction reflects real socioeconomic and financial differences relevant to credit risk modeling. Younger individuals often lack established credit histories and income stability, whereas older individuals tend to have more financial experience and responsibilities, which can affect their credit behavior in meaningful ways.
Moreover, in our experiments, we applied the following modifications:
\begin{enumerate}
    \item The original labeling assigns $Y=1$ to individuals who experienced financial distress, which is negatively framed. We flipped the labels so that $Y=1$ corresponds to a positive outcome—no financial distress within two years.
    \item We pre-processed the dataset following the procedure of \citet{detoni2024personalized} 
    to ensure recourse tractability.
    \item The original dataset is highly imbalanced, with 139,974 non-distressed cases and only 10,026 distressed ones. To avoid skewed classifiers and unreliable conclusions, we mitigate the imbalance by downsampling the majority class, resulting in an even 50/50 split between distressed and non-distressed individuals.
\end{enumerate}

\paragraph{German Credit.} It collects information about several individuals from a German bank's data from 1994.
It contains details about the socioeconomic situation of individuals, such as their employment, housing, savings, \textit{etc.}
Besides, the set of features includes some sensitive information as well, such as gender and age.
In this classification task, the objective is to predict whether an individual should obtain a good or bad credit score.
This dataset is considerably smaller than the previous ones, containing only 1,000 instances and 20 features, and it is publicly available in the UCI repository.
In our experiments, \textit{age} has been considered as the sensitive attribute, and we have divided the instances into two (sensitive) groups based on whether their age is greater than 30 or less than or equal to 30.
We pre-processed the dataset as detailed in \citet{pawelczyk1carla}.

\subsection{Fairness in Prediction}

We now describe the fairness-enhancing intervention selected to address fairness at the prediction level (Section~\ref{sec:experiments}).

\paragraph{Equality of opportunity by post-processing (\texttt{POSTPRO}).}
\citet{hardt2016equality} propose a fairness-enhancing intervention that modifies the outcomes of a given predictor to satisfy a given fairness property defined by either \textit{equality of opportunity} or \textit{equality of odds}.
In the case of binary predictors, their method flips the decisions with a given probability to satisfy the fairness criteria.
The code for this method can be found in the AIF360\footnote{\url{https://aif360.readthedocs.io/en/stable/}} Python package.
We selected \texttt{POSTPRO} not only for its widespread use in fairness-in-prediction benchmarking, but also because it is compatible with any base classifier, enabling a fair comparison with \texttt{MISOB}.

\begin{table*}[b]
\caption{\textbf{Hyperparameters for each recourse method used in Section~\ref{sec:experiments}.} For CCHVAE, we denote with \textsc{\#mutable\_feat} the number of actionable features available in each dataset.
Please refer to the source code and to the implementation of \citet{pawelczyk1carla} for additional details. 
}
\label{tab:hyperparams}
\begin{center}
\begin{small}
\begin{sc}
\scalebox{0.99}{\begin{tabular}{llll}
\toprule
Method & Hyperparameter & Type & Value \\
\midrule
GS & n/a & n/a & n/a \\
\midrule
WT & feature\_cost & list & None \\
& lr & float & 0.01 \\
& lambda\_param & float & 0.01 \\
& n\_iter & int & 1000 \\
& t\_max\_min & float & 0.5 \\
& norm & int & 1 \\
& clamp & boolean & True \\
& loss\_type & string & ``MSE'' \\
& y\_target & list & [0,1] \\
& binary\_cat\_features & boolean & True \\
\midrule
CCHVAE & n\_search\_samples & int & 100 \\
& p\_norm & \{1,2\} & 1 \\
& step & float & 0.1 \\
& max\_iter & int & 1000 \\
& clamp & boolean & True \\
& binary\_cat\_features & boolean & True \\
& vae\_params & dict & \\
& - layers & list & [\#mutable\_feat, 64, 32, 8] \\
& - train & boolean & True \\
& - lambda\_reg & float & 1e-6 \\
& - epoch & int & 5 \\
& - lr & float & 1e-3 \\
& - batch\_size & int & 128 \\
\bottomrule
\end{tabular}}
\end{sc}
\end{small}
\end{center}
\end{table*}

\subsection{Recourse Methods}

Here, we describe in greater detail the recourse methods used in our experimental evaluation (Section~\ref{sec:experiments}).
For all the recourse methods, we consider the default hyperparameter settings (see Table~\ref{tab:hyperparams}) and the implementation of \citet{pawelczyk1carla}.

\begin{itemize}
    \item \textbf{Growing Spheres (GS) \cite{laugel2017inverse}}: it is a random search algorithm, which generates samples around the instance under consideration using growing hyperspheres until a point with the positive class label is found;
    \item \textbf{Differentiable Recourse (WT)\cite{wachter2017counterfactual}}: it is a method that generates recommendations by minimizing an objective function using gradient descent to find a nearby ``possible world'' in which the individual receives a positive outcome, measuring proximity using the $\ell_1$ norm;
    \item \textbf{Counterfactual Conditional Heterogeneous Variational Autoencoder (CCHVAE) \cite{pawelczyk2020learning}}: it generates recommendations by ensuring that the latter maps the instance into a region of high data density that is as close as possible to the original instance, where proximity is computed in a lower-dimensional latent space defined by the VAE. 
\end{itemize}

\subsection{Additional Training and Implementation Details} 

We employed mini-batches both in the regular training and iterative procedures for both NN and linear regression (LR) to mitigate overfitting during training.
Specifically, we used a batch size of 256, a learning rate of 0.001, and the Adam optimizer.
The base classifiers were trained for 6 epochs.
In the case of \texttt{POSTPRO}, we performed the fairness-enhancing intervention after this training phase.
For the \texttt{MISOB} framework, we implemented a two-stage training strategy: a warm-up phase of 3 epochs, followed by 3 additional epochs of iterative optimization.
We experimented with longer training schedules but did not observe substantial performance gains.
Therefore, we opted to limit the number of epochs to maintain training efficiency.
Lastly, we have run all our experiments in a server with eight NVIDIA A5000 GPUs with an AMD EPYC 7252 CPU and 377GB of RAM.
Nevertheless, our approach is neither RAM nor CPU/GPU intensive.

\section{Further Results on Additional Classification Tasks}
\label{app:additional_results}

Table \ref{tab:results_german_age} and \ref{tab:results_giveme} present additional experimental results using the \textsc{German Credit} and \textsc{GiveMeSomeCredit} datasets, respectively.
In both cases, we consider \textit{age} as the sensitive attribute.
Alongside a feed-forward neural network (NN), we also evaluate a simple \textit{linear regression} (LR) classifier as an additional base model.

The results across these datasets and base classifiers reflect the same trends observed in the experiments from the main text with the \textsc{Adult} dataset.
Specifically, \texttt{MISOB} consistently achieves the best worst-group performance, outperforming all other methods in terms of minimax fairness guarantees, both across the entire recourse pipeline and in the predictions of the base classifier, while incurring minimal or no loss in overall accuracy.
Notably, although \texttt{POSTPRO} reduces TPR disparity at the base classifier level, \textit{it does so by lowering the TPR of the privileged group rather than improving that of the unprivileged group.}
As a result, the reduction in TPR gap comes at the cost of a higher burden and cost, and in some cases, reduced base accuracy.

These additional results also offer further insights. 
\texttt{MISOB} remains effective even on smaller datasets. 
However, we observe a higher standard deviation in performance metrics across all methods, a phenomenon also noted in the fairness literature \cite{barrainkua2024uncertainty}, which can complicate the extraction of reliable conclusions. Finally, especially in the case of the \textsc{GiveMeSomeCredit} dataset, we note that low data quality may hinder the training of strong base classifiers. Nevertheless, \texttt{MISOB} is still able to significantly reduce burden and cost, even when the base classifier itself performs poorly. 

\begin{table*}[t]
\caption{\textbf{Empirical average test results on the \textsc{German credit} dataset across different base classifiers and recourse methods.}
The best results for each setting are highlighted in bold, and we report whether the target metric should increase (${\uparrow}$) or decrease (${\downarrow}$).
We set \textit{age} as the sensitive attribute.
For \texttt{MISOB}, we set $\alpha=0.3$ in all experiments (\textit{cf.} Eqn.~\ref{eq:weighting}).
Each cell shows the average and standard deviation over 10 runs. 
%
}
\label{tab:results_german_age}
\begin{center}
\begin{small}
\begin{sc}
\scalebox{0.8}{\begin{tabular}{cccccccccccc}
\toprule
Base & Recourse & Fairness & acc  & \multicolumn{2}{c}{Burden (Eq.~\ref{eq:burden_eop})} & \multicolumn{2}{c}{TPR} & \multicolumn{2}{c}{Cost (Eq.~\ref{eq:cost_group})} & \multicolumn{2}{c}{AR}  \\
Model & method & strategy & overall ${\uparrow}$ & worst ${\downarrow}$ & $\Delta$ ${|\downarrow|}$ & worst ${\uparrow}$ & $\Delta$ ${|\downarrow|}$ & worst ${\downarrow}$ & $\Delta$ ${|\downarrow|}$ & worst ${\uparrow}$ & $\Delta$ ${|\downarrow|}$  \\
\midrule
NN & GS & - & \textbf{0.62}$_{\pm 0.13}$ & 30.92$_{\pm 61.27}$ & 5.07$_{\pm 57.24}$ & 0.75$_{\pm 0.29}$ & 0.02$_{\pm 0.33}$ & 44.43$_{\pm 75.91}$ & 7.89$_{\pm 89.73}$ & 0.74$_{\pm 0.26}$ & \textbf{0.02}$_{\pm 0.30}$ \\
NN & GS & \texttt{POSTPRO} & 0.61$_{\pm 0.03}$ &  51.46$_{\pm 42.93}$ & 27.59$_{\pm 31.75}$ &  0.05$_{\pm 0.09}$ &  \textbf{0.01}$_{\pm 0.12}$ &  95.31$_{\pm 7.06}$ &  8.73$_{\pm 10.26}$ &  0.04$_{\pm 0.07}$ &  0.02$_{\pm 0.12}$ \\
NN & GS & \texttt{MISOB} & 0.61$_{\pm 0.12}$ & \textbf{13.78}$_{\pm 39.10}$ & \textbf{1.36}$_{\pm 32.41}$ & \textbf{0.76}$_{\pm 0.29}$ & 0.04$_{\pm 0.28}$ & \textbf{20.98}$_{\pm 46.01}$ & \textbf{3.15}$_{\pm 48.36}$ &  \textbf{0.75}$_{\pm 0.26}$ & 0.06$_{\pm 0.30}$  \\
\midrule
NN & WT & - & \textbf{0.62}$_{\pm 0.13}$ &  15.69$_{\pm 44.71}$ & 6.84$_{\pm 24.71}$ & 0.75$_{\pm 0.29}$ & 0.02$_{\pm 0.33}$ & 20.79$_{\pm 58.15}$ & 7.33$_{\pm 35.81}$ & 0.74$_{\pm 0.26}$ & \textbf{0.02}$_{\pm 0.30}$ \\
NN & WT & \texttt{POSTPRO} & 0.61$_{\pm 0.03}$ &  27.57$_{\pm 31.55}$ & 2.82$_{\pm 29.49}$ &  0.05$_{\pm 0.09}$ &  \textbf{0.01}$_{\pm 0.12}$ &  48.40$_{\pm 36.80}$ & 3.41$_{\pm 24.57}$ &  0.04$_{\pm 0.07}$ &  \textbf{0.02}$_{\pm 0.12}$ \\
NN & WT & \texttt{MISOB} & \textbf{0.62}$_{\pm 0.12}$ & \textbf{1.07}$_{\pm 24.71}$ & \textbf{0.73}$_{\pm 44.71}$ & \textbf{0.84}$_{\pm 0.32}$ & 0.02$_{\pm 0.12}$ & \textbf{1.33}$_{\pm 1.06}$ & \textbf{0.47}$_{\pm 2.77}$ & \textbf{0.84}$_{\pm 0.17}$ & \textbf{0.02}$_{\pm 0.19}$ \\
\midrule
NN & CCHVAE & - & \textbf{0.62}$_{\pm 0.13}$ & 21.28$_{\pm 30.30}$ & \textbf{2.98}$_{\pm 28.81}$ & 0.75$_{\pm 0.29}$ & 0.02$_{\pm 0.33}$ & 30.24$_{\pm 39.28}$ & 3.45$_{\pm 39.27}$ & 0.74$_{\pm 0.26}$ & \textbf{0.02}$_{\pm 0.30}$ \\
NN & CCHVAE & \texttt{POSTPRO} & 0.61$_{\pm 0.03}$ &  71.10$_{\pm 31.93}$ &  3.72$_{\pm 7.92}$ &  0.05$_{\pm 0.09}$ &  \textbf{0.01}$_{\pm 0.12}$ & 81.10$_{\pm 44.04}$ &  \textbf{1.51}$_{\pm 3.07}$ &  0.04$_{\pm 0.07}$ &  \textbf{0.02}$_{\pm 0.12}$ \\
NN & CCHVAE & \texttt{MISOB} & \textbf{0.62}$_{\pm 0.11}$ & \textbf{9.57}$_{\pm 24.46}$ & 3.89$_{\pm 24.48}$ & \textbf{0.83}$_{\pm 0.22}$ & 0.08$_{\pm 0.11}$ & \textbf{9.78}$_{\pm 32.68}$ & 8.05$_{\pm 15.15}$ & \textbf{0.84}$_{\pm 0.21}$ & 0.06$_{\pm 0.11}$ \\
\midrule
LR & GS & - & \textbf{0.65}$_{\pm 0.10}$ & 13.67$_{\pm 7.13}$ & 9.68$_{\pm 2.60}$ & 0.77$_{\pm 0.27}$ & \textbf{0.01}$_{\pm 0.23}$ & 19.66$_{\pm 4.36}$ & 12.84$_{\pm 9.65}$ & 0.75$_{\pm 0.22}$ & \textbf{0.02}$_{\pm 0.23}$  \\
LR & GS & \texttt{POSTPRO} & 0.63$_{\pm 0.05}$ &  32.18$_{\pm 24.37}$ & \textbf{3.91}$_{\pm 16.55}$ &  0.08$_{\pm 0.08}$ &  0.02$_{\pm 0.07}$ & 73.67$_{\pm 42.41}$ &  10.15$_{\pm 27.05}$ &  0.05$_{\pm 0.02}$ &  \textbf{0.02}$_{\pm 0.07}$ \\
LR & GS & \texttt{MISOB} & 0.63$_{\pm 0.06}$ & \textbf{5.09}$_{\pm 5.16}$ & 3.97$_{\pm 1.12}$ & \textbf{0.80}$_{\pm 0.27}$ & 0.05$_{\pm 0.23}$ & \textbf{7.69}$_{\pm 4.36}$ & \textbf{5.28}$_{\pm 9.65}$ & \textbf{0.82}$_{\pm 0.17}$ & \textbf{0.02}$_{\pm 0.09}$  \\
\midrule
LR & WT & - & 0.65$_{\pm 0.10}$ & 1.14$_{\pm 2.39}$ & 0.45$_{\pm 1.21}$ &  0.77$_{\pm 0.27}$ & \textbf{0.01}$_{\pm 0.23}$ & 1.70$_{\pm 2.72}$ & 0.64$_{\pm 1.81}$ & 0.75$_{\pm 0.22}$ & \textbf{0.02}$_{\pm 0.23}$ \\
LR & WT & \texttt{POSTPRO} & 0.63$_{\pm 0.05}$ &  5.83$_{\pm 6.90}$ & 3.58$_{\pm 2.84}$ &  0.08$_{\pm 0.08}$ &  0.02$_{\pm 0.07}$ &  10.15$_{\pm 12.59}$ &  7.25$_{\pm 3.59}$ &  0.05$_{\pm 0.02}$ &  \textbf{0.02}$_{\pm 0.07}$ \\
LR & WT & \texttt{MISOB} & \textbf{0.67}$_{\pm 0.05}$ &  \textbf{0.11}$_{\pm 0.32}$ & \textbf{0.11}$_{\pm 0.32}$ & \textbf{0.92}$_{\pm 0.12}$ & \textbf{0.01}$_{\pm 0.11}$ & \textbf{0.11}$_{\pm 0.32}$ & \textbf{0.11}$_{\pm 0.32}$ & \textbf{0.91}$_{\pm 0.10}$ & \textbf{0.02}$_{\pm 0.11}$  \\
\midrule
LR & CCHVAE & - & 0.65$_{\pm 0.10}$ & 15.68$_{\pm 22.58}$ & 4.30$_{\pm 19.30}$ &  0.77$_{\pm 0.27}$ & \textbf{0.01}$_{\pm 0.23}$ & 27.28$_{\pm 39.48}$ & \textbf{4.03}$_{\pm 32.76}$ & 0.75$_{\pm 0.22}$ & \textbf{0.02}$_{\pm 0.23}$ \\
LR & CCHVAE & \texttt{POSTPRO} & 0.63$_{\pm 0.05}$ &  56.63$_{\pm 26.03}$ &  3.31$_{\pm 35.21}$ &  0.08$_{\pm 0.08}$ &  0.02$_{\pm 0.07}$ & 96.10$_{\pm 56.92}$ & 5.58$_{\pm 19.81}$ &  0.05$_{\pm 0.02}$ &  \textbf{0.02}$_{\pm 0.07}$ \\
LR & CCHVAE & \texttt{MISOB} & \textbf{0.66}$_{\pm 0.09}$ & \textbf{4.50}$_{\pm 16.61}$ & \textbf{3.02}$_{\pm 2.81}$ & \textbf{0.84}$_{\pm 0.23}$ & 0.05$_{\pm 0.28}$ & \textbf{12.52}$_{\pm 34.00}$ & 8.87$_{\pm 28.04}$ &  \textbf{0.86}$_{\pm 22.67}$ & 0.05$_{\pm 16.57}$  \\
\bottomrule
\end{tabular}}
\end{sc}
\end{small}
\end{center}
\end{table*}

\begin{table*}[t]
\caption{\textbf{Empirical average test results on the \textsc{GiveMeSomeCredit} dataset across different base classifiers and recourse methods.}
The best results for each setting are highlighted in bold, and we report whether the target metric should increase (${\uparrow}$) or decrease (${\downarrow}$).
We set \textit{age} as the sensitive attribute.
For \texttt{MISOB}, we set $\alpha=0.3$ in all experiments (\textit{cf.} Eqn.~\ref{eq:weighting}).
Each cell shows the average and standard deviation over 10 runs. }
\label{tab:results_giveme}
\begin{center}
\begin{small}
\begin{sc}
\scalebox{0.8}{\begin{tabular}{cccccccccccc}
\toprule
Base & Recourse & Fairness & acc  & \multicolumn{2}{c}{Burden (Eq.~\ref{eq:burden_eop})} & \multicolumn{2}{c}{TPR} & \multicolumn{2}{c}{Cost (Eq.~\ref{eq:cost_group})} & \multicolumn{2}{c}{AR}  \\
Model & Method & strategy & overall ${\uparrow}$ & worst ${\downarrow}$ & $\Delta$ ${|\downarrow|}$ & worst ${\uparrow}$ & $\Delta$ ${|\downarrow|}$ & worst ${\downarrow}$ & $\Delta$ ${|\downarrow|}$ & worst ${\uparrow}$ & $\Delta$ ${|\downarrow|}$  \\
\midrule
NN & GS & - & \textbf{0.56}$_{\pm 0.02}$ &  3.27$_{\pm 0.88}$ & \textbf{0.20}$_{\pm 0.56}$ &  0.30$_{\pm 0.11}$ & 0.33$_{\pm 0.12}$ & 25.04$_{\pm 7.29}$ &  3.79$_{\pm 2.37}$ &  0.24$_{\pm 0.09}$ &  0.26$_{\pm 0.10}$ \\
NN & GS & \texttt{POSTPRO} & 0.53$_{\pm 0.01}$ &  5.32$_{\pm 3.08}$ & 3.79$_{\pm 1.11}$ & 0.25$_{\pm 0.18}$ &  \textbf{0.01}$_{\pm 0.04}$ &  23.40$_{\pm 7.51}$ &  2.89$_{\pm 7.62}$ &  0.32$_{\pm 0.40}$ &  \textbf{0.13}$_{\pm 0.10}$ \\
NN & GS & \texttt{MISOB} & \textbf{0.56}$_{\pm 0.03}$ &  \textbf{1.65}$_{\pm 1.82}$ &  \textbf{0.49}$_{\pm 1.00}$ &  0.38$_{\pm 0.24}$ & 0.37$_{\pm 0.27}$ & \textbf{10.30}$_{\pm 6.17}$ & \textbf{0.01}$_{\pm 4.44}$ & \textbf{0.33}$_{\pm 0.22}$ & 0.34$_{\pm 0.23}$ \\
\midrule
NN & WT & - & 0.56$_{\pm 0.02}$ & 1.78$_{\pm 0.71}$ & 0.50$_{\pm 0.51}$ &  0.30$_{\pm 0.11}$ & 0.33$_{\pm 0.12}$ &  14.39$_{\pm 2.49}$ & 6.24$_{\pm 1.92}$ & 0.24$_{\pm 0.09}$ &  0.26$_{\pm 0.10}$ \\
NN & WT & \texttt{POSTPRO} & 0.53$_{\pm 0.01}$ &  2.55$_{\pm 2.17}$ & 0.84$_{\pm 4.97}$ &  0.25$_{\pm 0.18}$ &   \textbf{0.01}$_{\pm 0.04}$ &  13.62$_{\pm 4.73}$ &  5.79$_{\pm 7.06}$ &  0.32$_{\pm 0.40}$ &  \textbf{0.13}$_{\pm 0.10}$ \\
NN & WT & \texttt{MISOB} & \textbf{0.59}$_{\pm 0.05}$ &  \textbf{1.32}$_{\pm 1.64}$ &  \textbf{0.31}$_{\pm 1.19}$ &  \textbf{0.55}$_{\pm 0.13}$ &  0.24$_{\pm 0.11}$ &  \textbf{11.37}$_{\pm 6.71}$ &  \textbf{4.47}$_{\pm 4.25}$ & \textbf{0.46}$_{\pm 0.10}$ & 0.20$_{\pm 0.16}$ \\
\midrule
NN & CCHVAE & - & 0.56$_{\pm 0.02}$ & 6.43$_{\pm 3.48}$ & 1.04$_{\pm 3.43}$ & 0.30$_{\pm 0.11}$ & 0.33$_{\pm 0.12}$ &  35.00$_{\pm 13.03}$ & 1.49$_{\pm 17.43}$ &  0.24$_{\pm 0.09}$ &  0.26$_{\pm 0.10}$ \\
NN & CCHVAE & \texttt{POSTPRO} & 0.53$_{\pm 0.01}$ &  20.62$_{\pm 24.14}$ &  10.17$_{\pm 17.64}$ &  0.25$_{\pm 0.18}$ &   \textbf{0.01}$_{\pm 0.04}$ &  29.21$_{\pm 35.16}$ & 7.57$_{\pm 31.21}$ &  0.32$_{\pm 0.40}$ &  \textbf{0.13}$_{\pm 0.10}$ \\
NN & CCHVAE & \texttt{MISOB} & \textbf{0.57}$_{\pm 0.04}$ & \textbf{4.15}$_{\pm 2.40}$ & \textbf{0.76}$_{\pm 2.10}$ & \textbf{0.43}$_{\pm 0.16}$ & 0.31$_{\pm 0.13}$ & \textbf{24.25}$_{\pm 11.47}$ & \textbf{1.19}$_{\pm 9.52}$ & \textbf{0.36}$_{\pm 0.15}$ & 0.28$_{\pm 0.14}$ \\
\midrule
LR & GS & - & 0.51$_{\pm 0.09}$ &  10.95$_{\pm 12.15}$ & 5.91$_{\pm 1.98}$ &  0.17$_{\pm 0.25}$ & 0.03$_{\pm 0.05}$ & 33.38$_{\pm 26.30}$ &  13.63$_{\pm 8.47}$ & 0.16$_{\pm 0.18}$ & \textbf{0.03}$_{\pm 0.03}$ \\
LR & GS & \texttt{POSTPRO} & \textbf{0.54}$_{\pm 0.10}$ &  17.31$_{\pm 11.86}$ & 6.21$_{\pm 2.30}$ &  0.11$_{\pm 0.23}$ &   \textbf{0.01}$_{\pm 0.03}$ &  39.23$_{\pm 30.92}$ &  6.55$_{\pm 6.33}$ &  0.09$_{\pm 0.18}$ &  0.04$_{\pm 0.02}$ \\
LR & GS & \texttt{MISOB} & 0.52$_{\pm 0.02}$ & \textbf{2.59}$_{\pm 1.57}$ &  \textbf{1.51}$_{\pm 1.32}$ & \textbf{0.45}$_{\pm 0.24}$ &  0.04$_{\pm 0.15}$ & \textbf{16.61}$_{\pm 0.04}$ &  \textbf{4.60}$_{\pm 1.64}$ & \textbf{0.34}$_{\pm 0.36}$ & 0.06$_{\pm 0.17}$ \\
\midrule
LR & WT & - & 0.51$_{\pm 0.09}$ &  2.75$_{\pm 2.12}$ & 0.77$_{\pm 0.87}$ & 0.17$_{\pm 0.25}$ & 0.03$_{\pm 0.05}$ & 10.63$_{\pm 6.29}$ & 4.74$_{\pm 1.63}$ &  0.16$_{\pm 0.18}$ &  0.03$_{\pm 0.03}$ \\
LR & WT & \texttt{POSTPRO} & 0.54$_{\pm 0.10}$ & 5.25$_{\pm 1.43}$ & \textbf{0.89}$_{\pm 0.76}$ &  0.11$_{\pm 0.23}$ &   \textbf{0.01}$_{\pm 0.03}$ &  16.16$_{\pm 4.44}$ &  7.93$_{\pm 2.48}$ &  0.09$_{\pm 0.18}$ &  0.04$_{\pm 0.02}$ \\
LR & WT & \texttt{MISOB} & \textbf{0.56}$_{\pm 0.11}$ &  \textbf{1.63}$_{\pm 0.30}$ &  0.90$_{\pm 0.57}$ & \textbf{0.54}$_{\pm 0.28}$ & 0.04$_{\pm 0.35}$ & \textbf{6.67}$_{\pm 6.28}$ & \textbf{3.11}$_{\pm 1.59}$ & \textbf{0.51}$_{\pm 0.27}$ & \textbf{0.01}$_{\pm 0.36}$ \\
\midrule
LR & CCHVAE & - & 0.51$_{\pm 0.09}$ & 14.81$_{\pm 11.34}$ & \textbf{3.26}$_{\pm 10.71}$ & 0.17$_{\pm 0.25}$ & 0.03$_{\pm 0.05}$ & 48.08$_{\pm 17.34}$ &  7.77$_{\pm 35.51}$ &  0.16$_{\pm 0.18}$ &  \textbf{0.03}$_{\pm 0.03}$ \\
LR & CCHVAE & \texttt{POSTPRO} & \textbf{0.54}$_{\pm 0.10}$ &  17.51$_{\pm 25.99}$ &  6.30$_{\pm 16.01}$ &  0.11$_{\pm 0.23}$ &   \textbf{0.01}$_{\pm 0.03}$ &  56.87$_{\pm 39.04}$ & \textbf{3.33}$_{\pm 28.95}$ &  0.09$_{\pm 0.18}$ &  0.04$_{\pm 0.02}$ \\
LR & CCHVAE & \texttt{MISOB} & \textbf{0.54}$_{\pm 0.07}$ & \textbf{7.55}$_{\pm 5.73}$ &  5.50$_{\pm 10.98}$ & \textbf{0.44}$_{\pm 0.38}$ &  \textbf{0.01}$_{\pm 0.39}$ & \textbf{29.61}$_{\pm 34.15}$ &  8.61$_{\pm 37.65}$ & \textbf{0.44}$_{\pm 0.36}$ &  0.04$_{\pm 0.37}$ \\
\bottomrule
\end{tabular}}
\end{sc}
\end{small}
\end{center}
\end{table*}

\end{document}